\documentclass[10pt,twocolumn,letterpaper]{article}

\usepackage[pagenumbers]{cvpr} 

\usepackage[dvipsnames]{xcolor}

\definecolor{cvprblue}{rgb}{0.21,0.49,0.74}
\usepackage[pagebackref,breaklinks,colorlinks,citecolor=cvprblue]{hyperref}
\usepackage{color, colortbl}
\usepackage[dvipsnames]{xcolor}
\usepackage{multirow}
\usepackage{makecell}
\usepackage{pifont}
\usepackage{booktabs}
\usepackage{algorithm}
\usepackage{algorithmicx}
\usepackage{algpseudocode}
\usepackage{eqparbox}
\usepackage{graphicx}

\newcommand{\cmark}{\ding{51}}
\newcommand{\xmark}{\ding{55}}
\definecolor{baselinecolor}{rgb}{0.9, 0.9, 1.}
\definecolor{graycolor}{gray}{0.9}
\newcommand{\baseline}[1]{\cellcolor{baselinecolor}{#1}}
\newcommand{\graybase}[1]{\cellcolor{graycolor}{#1}}

\begin{document}

\title{\makecell[c]{PAD: Self-Supervised Pre-Training with Patchwise-Scale Adapter \\ for Infrared Images}}

\author{Tao Zhang$^{1,2}$\quad Kun Ding$^{1,3}$\quad Jinyong Wen$^{1,2}$\quad Yu Xiong$^{1,2}$\quad Zeyu Zhang$^{3}$\\ Shiming Xiang$^{1,2}$\quad Chunhong Pan$^{1,3}$\\
\small $^1$ State Key Laboratory of Multimodal Artificial Intelligence Systems, Institute of Automation, 
CAS, Beijing, China\\
\small $^2$ School of Artificial Intelligence, University of Chinese Academy of Sciences, Beijing, China \\
\small $^3$ Research Center of Aerospace Information, Institute of Automation, 
CAS, Beijing, China\\
{\tt\small \{zhangtao2021, kun.ding\}@ia.ac.cn} \quad {\tt\small \{smxiang, chpan\}@nlpr.ia.ac.cn}
}

\maketitle
\begin{abstract}
Self-supervised learning (SSL) for RGB images has achieved significant success, yet there is still limited research on SSL for infrared images, primarily due to three prominent challenges: 1) the lack of a suitable large-scale infrared pre-training dataset, 2) the distinctiveness of non-iconic infrared images rendering common pre-training tasks like masked image modeling (MIM) less effective, and 3) the scarcity of fine-grained textures making it particularly challenging to learn general image features. To address these issues, we construct a \textbf{M}ulti-\textbf{S}cene \textbf{I}nfrared \textbf{P}re-training (MSIP) dataset comprising 178,756 images, and introduce object-sensitive random RoI cropping, an image preprocessing method, to tackle the challenge posed by non-iconic images. To alleviate the impact of weak textures on feature learning, we propose a pre-training paradigm called \textbf{P}re-training with \textbf{AD}apter (PAD), which uses adapters to learn domain-specific features while freezing parameters pre-trained on ImageNet to retain the general feature extraction capability. This new paradigm is applicable to any transformer-based SSL method. Furthermore, to achieve more flexible coordination between pre-trained and newly-learned features in different layers and patches, a patchwise-scale adapter with dynamically learnable scale factors is introduced. Extensive experiments on three downstream tasks show that PAD, with only 1.23M pre-trainable parameters, outperforms other baseline paradigms including continual full pre-training on MSIP. Our code and dataset are available at \url{https://github.com/casiatao/PAD}.
\end{abstract}    

\section{Introduction}
\label{sec:intro}

Inspired by the masked language modeling (MLM) \cite{kenton2019bert, radford2018improving, brown2020language} in natural language processing (NLP), various masked image modeling (MIM) methods \cite{chen2020generative, bao2021beit, mae} emerge and expand their scope beyond RGB images to other domains, including depth maps  \cite{bachmann2022multimae}, point clouds \cite{chen2023pimae}, and remote sensing images \cite{cong2022satmae}. However, there is a noticeable absence of SSL methods for infrared imagery. Infrared images, capturing thermal radiation, hold vast potential across domains such as surveillance \cite{bondi2020birdsai}, autonomous driving \cite{leira2021object}, and medical diagnostics \cite{chen2023near}. Compared to the RGB domain, annotated infrared images are even scarcer, severely limiting the development of computer vision in this area. Therefore, it is urgent to bridge the current research gap.

\begin{figure}[t]
  \vspace{-2mm}
  \centering
  \includegraphics[width=1.0\linewidth]{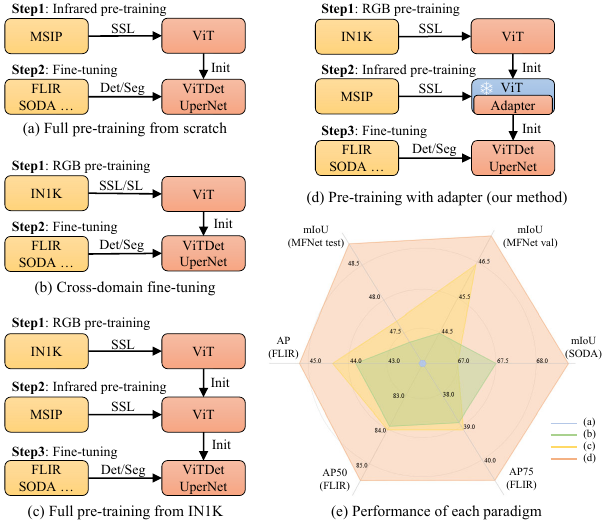}
  \vspace{-6mm}
  \caption{\textbf{The baseline paradigms (a-c) and PAD paradigm (d) of pre-training and fine-tuning for infrared images, along with their performance (e).} PAD achieves state-of-the-art performance on various downstream tasks compared to other paradigms.} 
  \label{fig:paradigm}
  \vspace{-6mm}
\end{figure}

\begin{figure}[t]
  \vspace{-2mm}
  \centering
  \includegraphics[width=1.0\linewidth]{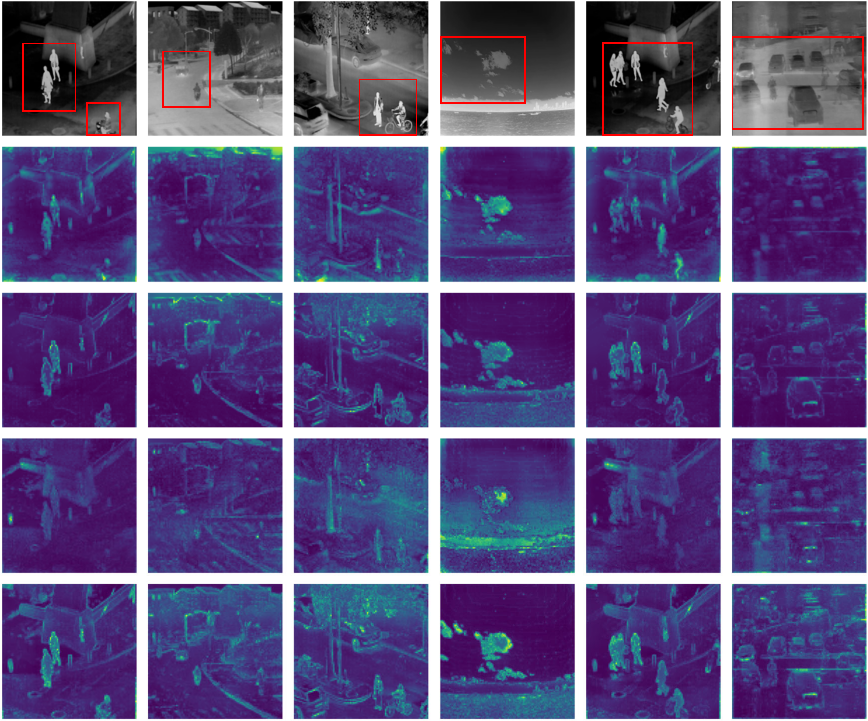}
  \vspace{-6mm}
  \caption{\textbf{Visualizations of original images and attention maps of the last self-attention layer in different paradigms based on MAE.} From top to bottom, each row represents: 1) original images from MSIP with red boxes indicating areas of interest, attention maps of 2) full pre-training from scratch, 3) cross-domain fine-tuning, 4) full pre-training from IN1K, 5) PAD.} 
  \label{fig:attention}
  \vspace{-6mm}
\end{figure}

However, there are three primary distinctions between infrared and RGB pre-training that make infrared SSL more challenging. Firstly, the quantity and scale of publicly available infrared datasets are relatively limited, lacking a large-scale pre-training dataset encompassing diverse application scenarios. Typically, most detection \cite{FLIR, ship} and segmentation \cite{soda, ha2017mfnet} datasets consist of only a few thousand or even a few hundred images. While some tracking datasets \cite{liu2020lsotb, li2021lasher} include several hundred thousand images, they are composed of multiple consecutive video sequences, introducing a notable redundancy for image-level SSL. To tackle this challenge, we construct a comprehensive unlabeled dataset, named Multi-Scene Infrared Pre-training (MSIP), consisting of 178,756 infrared images sourced from eight distinct infrared datasets \cite{dronevehicle, scut, jia2021llvip, aerial, ship, security, liu2020lsotb, li2021lasher}, covering a wide spectrum of scenarios.

Secondly, despite our efforts to compile a sizable dataset, the MSIP dataset significantly differs from ImageNet \cite{deng2009imagenet}, which has been meticulously curated to primarily feature single-object-centric images. In contrast, most infrared images depict outdoor scenes with multiple objects, including relatively small ones. While MIM-based methods \cite{wei2022maskfeat, mae} are more robust for images with multiple instances compared to contrastive learning approaches \cite{byol, mocov3}, the large variability in image scales within MSIP can lead to small objects being split into just a few patches. When subjected to a high mask ratio, these patches may either be fully masked or only leave a tiny portion visible. Consequently, a majority of visible patches consist of textureless background regions, increasing the complexity of pre-training and challenging the model's ability to extract valuable information. Additionally, objects occupying only a few patches are mainly attended by nearby patches, limiting the utilization of the long-range modeling capability of the transformer \cite{vaswani2017transformer}. Therefore, we propose an image preprocessing method termed \textit{random RoI cropping}, which employs selective search \cite{uijlings2013selectivesearch} along with random translation and scaling to crop images. This approach produces images with a closer resemblance to those in ImageNet, effectively mitigating the issue of small objects being fully masked.

The third distinction is the lack of fine-grained details in infrared images. A straightforward method of extending SSL methods from visible to infrared domain involves initializing the model with parameters pre-trained on ImageNet, followed by further pre-training on infrared data, as illustrated in \cref{fig:paradigm} (c). However, continual pre-training on infrared images tends to hurt the original model's feature extraction capability (the blurred edges in the 4th row of \cref{fig:attention}) and thereby damages the performance on downstream tasks, making it even worse than pre-training on ImageNet only (\cref{fig:paradigm} (b)). Similar phenomena have also been observed in NLP\textemdash models overly focus on domain-specific knowledge, disregarding general knowledge and resulting in unnatural text generation \cite{lawgpt, wu2023bloomberggpt}.

Motivated by recent trends in NLP \cite{hu2021lora} and advancements in visual tuning \cite{chen2022adaptformer}, we introduce a self-supervised pre-training paradigm, named Pre-training with Adapter (PAD), to address the issue of overfitting when performing pre-training in data-scarce domains like infrared. Specifically, as depicted in \cref{fig:paradigm} (d), the model is firstly pre-trained on ImageNet to acquire the general feature extraction ability and then pre-trained on MSIP. During infrared pre-training, the model parameters remain fixed while adapters are incorporated for training, enabling the learning of domain-specific feature extraction ability. It is noteworthy that PAD can be applied to any transformer-based SSL method. Furthermore, we introduce the \textit{patchwise-scale adapter}, which has dynamically learnable scale factors for different layers and patches. As such, the fusion between domain-specific and general features can be achieved in an adaptive manner, resulting in improved downstream performance without cumbersome manual adjustment of the scale factor. Results of extensive experiments on three downstream tasks \cite{soda, ha2017mfnet, FLIR}, summarized in \cref{fig:paradigm} (e), demonstrate the effectiveness of the proposed PAD paradigm.

Overall, the main contributions of this paper are:
\begin{itemize}
    \item We construct a large and diverse infrared pre-training dataset MSIP and provide extensive experimental benchmarks for the first time, filling the gap in infrared pre-training research.
    \item We introduce random RoI cropping to effectively tackle the issue associated with small objects for infrared SSL. 
    \item We propose the PAD paradigm, allowing the model to learn domain-specific features while retaining the ability to acquire general features. We also introduce the patchwise-scale adapter to adaptively determine the importance of domain-specific features for each patch.
    \item Extensive experiments demonstrate the effectiveness and strong generalization capability of our method.
\end{itemize}
\section{Related works}
\label{sec:related works}

\begin{figure*}[t]
  \vspace{-2mm}
  \centering
  \includegraphics[width=1.0\linewidth]{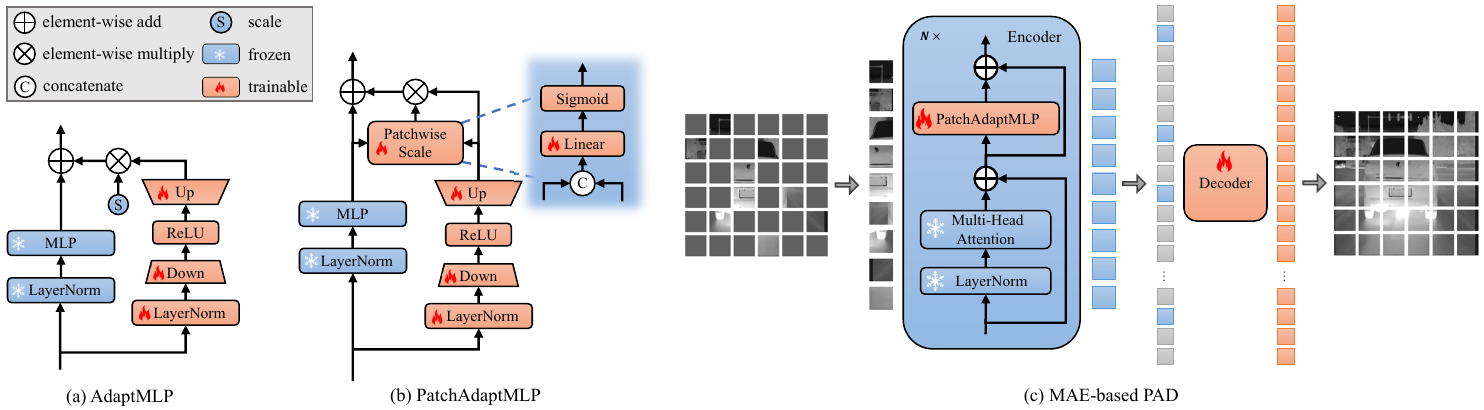}
  \vspace{-6mm}
  \caption{\textbf{Architecture of AdaptMLP (a), PatchAdaptMLP (b), and MAE-based PAD (c).} During infrared pre-training, parameters in the parallel branch and patchwise-scale module of PatchAdaptMLP are updated, while other parts of the encoder are frozen.} 
  \label{fig:method}
  \vspace{-6mm}
\end{figure*}

\noindent \textbf{Self-supervised learning} aims to enhance the feature representation ability of models by designing proxy tasks that leverage inherent features of data as supervisory signals. Contrastive learning \cite{byol, dino, mocov3} employs instance discrimination as a proxy task, whereas MIM-based methods \cite{bao2021beit, wei2022maskfeat, mae, xie2022simmim, xie2022mfm} corrupt images using diverse masks and formulate various reconstruction objectives as proxy tasks. SSTN \cite{munir2021sstn} leverages contrastive learning on RGBT image pairs for pre-training. To the best of our knowledge, we are the first to investigate MIM-based pre-training methods using only infrared images.

\noindent \textbf{Self-supervised learning on non-iconic datasets} focuses on addressing the issues of applying SSL on datasets like COCO \cite{lin2014coco}, which resembles realistic scenarios with multiple objects. Self-EMD \cite{liu2020self} pre-trains models by aligning local image features. UniVIP \cite{li2022univip} combines three hierarchical levels of contrastive learning. R-MAE \cite{nguyen2023r} employs regions, along with additional encoder and decoder, to facilitate the training of MAE \cite{mae}. Our proposed random RoI cropping method modifies the pre-training images without impacting the pre-training pipeline and architecture.

\noindent \textbf{Adapter-based methods} employ adapters to efficiently fine-tune pre-trained models for downstream tasks. LoRA \cite{hu2021lora} utilizes a low-rank adapter to fine-tune the self-attention layer of large-scale pre-trained models. AdaptFormer \cite{chen2022adaptformer} introduces a similar adapter to adjust the multi-layer perceptron (MLP) block within transformers. ViT-adapter\cite{chen2022vitadapter} is proposed to adapt ViT \cite{vit} to dense prediction tasks. Different from these methods, our proposed PAD paradigm applies adapters for self-supervised pre-training in the infrared domain. Moreover, we also propose an adaptive method to learn patchwise scale factors, which effectively addresses the dynamic coordination problem between domain-specific and general features.

\section{Approach}
\label{sec:approach}

\subsection{Baseline paradigms}

We summarize several straightforward paradigms of infrared pre-training and fine-tuning as baseline paradigms.

\noindent
\textbf{Full pre-training from scratch.} The model is randomly initialized, then pre-trained on infrared images, and finally fine-tuned on downstream tasks (\cref{fig:paradigm} (a)). In this paradigm, the pre-trained model, exposed solely to textureless infrared images, has limited capacity for general feature extraction, leading to an overemphasis on brightness information, as demonstrated in the 2nd row of \cref{fig:attention}.

\noindent
\textbf{Cross-domain fine-tuning.} The model pre-trained on ImageNet-1K (IN1K) \cite{deng2009imagenet} is directly fine-tuned on infrared downstream tasks without prior pre-training on MSIP (\cref{fig:paradigm} (b)). This paradigm enables the convenient use of off-the-shelf pre-trained models \cite{mocov3, mae} from the RGB domain, with strong general feature extraction capability, as depicted in the 3rd row of \cref{fig:attention}. However, devoid of exposure to infrared images, these models lack the ability to extract domain-specific features, potentially limiting their performance on downstream tasks.

\noindent
\textbf{Full pre-training from IN1K.} The model is initially pre-trained on IN1K \cite{deng2009imagenet} and then further pre-trained on the infrared dataset before fine-tuning on downstream tasks (\cref{fig:paradigm} (c)). Intuitively, the model pre-trained on both RGB and infrared data should possess a combination of general and domain-specific feature extraction capabilities. However, full pre-training on MSIP tends to harm the general feature extraction capability of the model pre-trained on IN1K, as illustrated in the 4th row of \cref{fig:attention}.

\subsection{Pre-training with adapter}
To balance the general and domain-specific feature extraction capabilities of the model, we introduce the \textit{pre-training with adapter} (PAD) paradigm. As shown in \cref{fig:paradigm} (d), the model initially pre-trained on IN1K continues its pre-training on MSIP. However, in contrast to the \textit{full pre-training from IN1K} paradigm, in this step, the adapter is introduced for training to specialize in extracting domain-specific features, while parameters of the original model remain frozen to preserve the general feature extraction ability. The resulting pre-trained model, including the adapter, is subsequently fine-tuned on downstream tasks.

We initially employ the adapter in AdaptFormer \cite{chen2022adaptformer}, which is originally introduced to efficiently adapt pre-trained ViT \cite{vit} models to various downstream tasks. AdaptFormer modifies the MLP block of the original ViT by adding a parallel branch, resulting in an adapted version of the MLP known as AdaptMLP, as illustrated in \cref{fig:method} (a). The parallel branch, \ie the adapter, includes a LayerNorm (LN) \cite{ba2016layernorm} layer\footnote{In \cite{chen2022adaptformer}, there is a discrepancy about the position of LayerNorm between the structure diagram and the source code. Here, we follow their source code, where two distinct LayerNorm layers are applied to each of the two branches.}, a down-projection layer with weights $\mathbf{W}_{\text{down}}\in \mathbb{R}^{d\times r}$ and biases $\mathbf{b}_{\text{down}}\in \mathbb{R}^{r}$, a non-linear activation function ReLU \cite{agarap2018relu}, and an up-projection layer with weights $\mathbf{W}_{\text{up}}\in \mathbb{R}^{r \times d}$ and biases $\mathbf{b}_{\text{up}}\in \mathbb{R}^{d}$, where $d$ and $r$ ($r\ll d$) signify the channel number of tokens in ViT and the middle dimension of the adapter, respectively.

Given input features $x$, the output of the original MLP branch is
\begin{equation}
    x_{\text{mlp}} = \text{MLP}(\text{LN}(x)).
\end{equation}

The output of the parallel branch is computed as follows:
\begin{equation}
    x_{\text{adapt}} = \mathbf{W}_{\text{up}} \cdot  \text{ReLU}(\mathbf{W}_{\text{down}} \cdot \text{LN}(x) + \mathbf{b}_{\text{down}}) + \mathbf{b}_{\text{up}}.
\end{equation}

Then $x_{\text{adapt}}$ is fused with $x_{\text{mlp}}$ using a scale factor $s$, resulting in the output of AdaptMLP as follows:
\begin{equation}
    x'=x_{\text{mlp}}+s\cdot x_{\text{adapt}}.
    \label{eq:add}
\end{equation}

During infrared pre-training, only the parameters of the added adapter are updated. In subsequent sections, we refer to the adapter in AdaptFormer as the \textit{vanilla adapter}.

\subsection{Limitations of adapters for pre-training}
\label{subsec:limitations}
Several adapter-based fine-tuning methods, such as LoRA \cite{hu2021lora} and AdaptFormer \cite{chen2022adaptformer}, involve a scale factor, which influences the adapter in two aspects.
During training, the scale factor becomes coupled with the hyperparameter \textit{learning rate}, influencing the learning speed of adapters. After training, the scale factor explicitly governs the relative importance of domain-specific features when fused with general features. The theoretical proofs and corresponding experimental results are given in \cref{app:scale factor}.

In AdaptFormer \cite{chen2022adaptformer}, the scale factor remains consistent across various target tasks and model layers. Nevertheless, different layers within the model are tasked with extracting features at distinct hierarchical levels \cite{vit}, hence the importance of domain-specific features should differ across these layers. Additionally, as indicated in \cite{yoogatedprompt}, different tasks may require varying degrees of domain-specific features for each layer. One straightforward approach to tackle this issue involves assigning specific scale factors to each layer within each task. However, in the context of this paper, due to the gap between pre-training tasks and downstream tasks, as well as potential discrepancies between pre-training data and downstream data, the scale obtained during pre-training may be ineffective for various downstream tasks. Opting for task-specific scale factors would require individual pre-training for each task, contradicting the initial goal of the unified pre-trained model adaptable to various tasks.

We argue that, whether in pre-training or downstream tasks, different patches within an image demand varying levels of domain-specific features, corresponding to different scale factors. In the context of infrared imagery, it is intuitive that regions with high thermal radiation, such as car engines, and areas with lower temperatures, like the ground, exhibit distinct needs for domain-specific features. If the model can learn the variability in scale factors between different patches during pre-training, its performance on downstream tasks should improve. Assigning different scale factors to patches can be conceptualized as a spatial attention mechanism \cite{woo2018cbam} in the context of feature fusion, and it is also a generalized form of both layerwise and task-wise scale factors.

\subsection{Patchwise-scale adapter}
To capture the variability in scale factors of different patches, we propose the \textit{patchwise-scale adapter}, incorporating a simple patchwise-scale (PS) module into the vanilla adapter (VA) to predict scale factors for each patch. As illustrated in \cref{fig:method} (b), the PS module consists of a bias-free linear layer with weight $\textbf{W}_\text{linear} \in \mathbb{R}^{2d\times 1}$ and a Sigmoid activation function. The AdaptMLP with the incorporated PS module is referred to as PatchAdaptMLP. 

The original MLP features $x_{\text{mlp}}$ and adapter features $x_{\text{adapt}}$ are concatenated along the channel dimension. Then, they are passed through the linear layer and sigmoid function to predict the scale factor $s$ in \cref{eq:add} for each patch:

\begin{equation}
    s=\text{Sigmoid}(\textbf{W}_\text{linear} \cdot [x_{\text{mlp}} \vert x_{\text{adapt}}]),
    \label{eq:ps}
\end{equation}
where $[\cdot \vert \cdot]$ denotes the concatenation operation. As $s$ depends on input features, it can control the fusion of domain-specific and general features in a fine-grained manner.

\begin{figure}[t]
  \vspace{-2mm}
  \centering
  \includegraphics[width=1.0\linewidth]{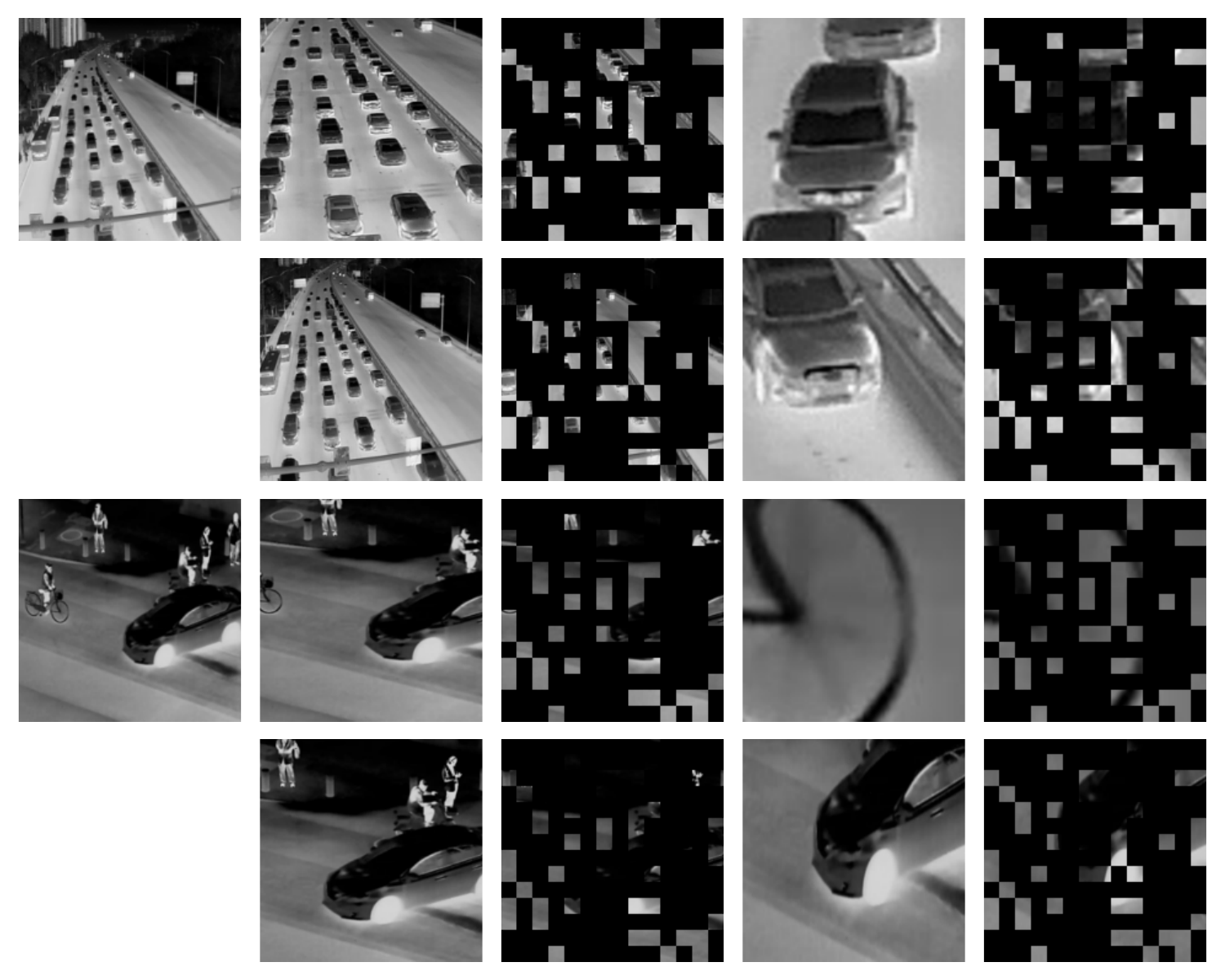}
  \vspace{-6mm}
  \caption{\textbf{Comparison of random resized cropping with our random RoI cropping.} From left to right, each column represents 1) original images from MSIP, 2) images generated by random resized cropping, 3) masked images of the second column, 4) images generated by random RoI cropping, and 5) masked images of the fourth column. Images generated by random RoI cropping are more similar to those in ImageNet, with the object centered.} 
  \label{fig:rrc}
  \vspace{-6mm}
\end{figure}

\subsection{Random RoI cropping}
Most MIM methods \cite{bao2021beit, mae} commonly employ random resized cropping for image preprocessing when pre-training on ImageNet. However, for non-iconic datasets, especially in the infrared domain where most images capture outdoor scenes with varying scales and multiple objects, small objects can be split into very few patches. When using random resized cropping with a high mask ratio, these patches may be entirely masked, as shown in the 3rd column of \cref{fig:rrc}, where many people and cars are almost fully masked. It is unreasonable to recover these completely occluded objects from the visible background patches. Moreover, objects that occupy only a few patches are mainly attended to by very nearby patches, making it difficult to leverage the long-range modeling capability of transformers \cite{vaswani2017transformer, vit}.

Therefore, we introduce an image preprocessing method termed \textit{random RoI cropping}. Initially, a vast number of regions of interest (RoIs) are generated for each pre-training image by the selective search method \cite{uijlings2013selectivesearch}. During each epoch of pre-training, the random RoI cropping randomly selects one of these candidate RoIs for each image. Subsequently, random translations within a defined range are applied to the center of the RoI, and its width and height are scaled within specified bounds, with constraints on the edge length and the ratio to the entire image area. The manipulated RoI is then utilized to crop the corresponding pre-training image, serving as the input for the pre-training pipeline. More detailed steps are in \cref{app:rrc}.

As shown in the 4th and 5th columns of \cref{fig:rrc}, this approach can crop out small objects within the images, resulting in cropped images that resemble object-centered ones in ImageNet and effectively addressing the issue of small targets being entirely masked.

\subsection{Architecture design}
The PatchAdaptMLP can serve as a plug-in structure applicable to any transformer-based SSL method. We primarily utilize MAE \cite{mae} as an infrared SSL framework, as depicted in \cref{fig:method} (c). Given an input image generated by random RoI cropping, it is partitioned into a series of patches with a large proportion being masked. The remaining visible patches are fed into the encoder, a ViT \cite{vit} with the original MLP replaced by the proposed PatchAdaptMLP. The encoded patches are concatenated with mask tokens and then passed into the decoder to predict the normalized pixel values for each patch. The loss function is the mean squared error between the actual and predicted pixel values.

\noindent
\textbf{Initialization of pre-training.} Following \cite{chen2022adaptformer}, the weights of the down projection layer in VA are initialized using Kaiming Normal \cite{he2015delving}. The weights of the up projection layer in VA and the linear layer in PS, along with the biases in VA, are initialized to zero. The remaining components of the model are initialized using weights pre-trained on IN1K with MAE. This initialization strategy ensures that the output of the adapter starts at zero, thus making the model perform consistently with the original one in the initial stages.

\noindent
\textbf{Initialization of fine-tuning.} The pre-trained encoder, with the patchwise-scale adapters, serves as the backbone when fine-tuning on downstream tasks.

\section{Experiments}
\label{sec:experiments}

\subsection{Experimental settings}

\begin{table*}[t]
    \centering
    \footnotesize
    \begin{tabular}{l c c c c|c|c c|c c c}
        \hline
        \multirow{2}{*}{Method} & \multirow{2}{*}{\text{\makecell[c]{Pre-training data}}} & \multirow{2}{*}{Epoch} & \multirow{2}{*}{\text{\makecell[c]{Total \\ params}}} & \multirow{2}{*}{\text{\makecell[c]{Training \\ params}}} & SODA & \multicolumn{2}{c|}{MFNet} & \multicolumn{3}{c}{FLIR} \\
         &  &  &  & & test & val & test & AP & AP50 & AP75 \\
         \hline
         \hspace{5pt} Random init & - & - &85.80M & - & 41.49 & 19.72 & 19.81 & 31.67 & 63.36 & 27.24 \\
         \hspace{5pt} EC-CNN$^\dagger$\cite{soda}  &  IN1K\cite{deng2009imagenet}  & - & - & - & 63.80 & - & - & - & - & - \\
         \hspace{5pt} SSTN101$^\dagger$\cite{munir2021sstn}  & FLIR\cite{munir2021sstn}  & 1000 & -  & -  & -  & - & - & - &  77.57 & - \\
         \hline
         \textcolor{gray}{{\makecell[l]{Full pre-training from scratch}}} & & & & & & & & & & \\
         \hspace{5pt} MAE \cite{mae}     & MSIP & 400  & 85.80M & 85.80M  & \graybase{61.46} & \graybase{39.58} & \graybase{42.98} & \graybase{39.71} & \graybase{76.33} &  \graybase{35.40} \\
         \hline
         \textcolor{gray}{{\makecell[l]{Cross-domain fine-tuning}}} & & & & & & & & & & \\
         \hspace{5pt} DeiT \cite{touvron2021deit}    & IN1K & 300  & 85.80M & - & \underline{68.51} & 43.94  & \underline{48.66} & 41.12 & 81.08 & 34.87\\
         \hspace{5pt} BeiT \cite{bao2021beit}    & IN1K & 300  & 85.70M & - & 64.09 & 42.42  & 45.05 & 36.26 & 71.48 & 31.49 \\
         \hspace{5pt} MaskFeat  \cite{wei2022maskfeat} & IN1K & 300  & 85.80M & - & 65.98 & 41.99  & 45.97 & 41.17 & 80.55 & 35.54 \\
         \hspace{5pt} MoCo v3 \cite{mocov3} & IN1K & 300  & 85.80M & - & 67.05 & 43.23  & 47.60 & 40.26 & 79.99 & 34.46 \\
         \hspace{5pt} MILAN  \cite{hou2022milan}  & IN1K & 400  & 85.80M & - & \textbf{68.93} & \underline{45.08} & 47.11 & 42.29 & 81.99 & 36.46\\
         \hspace{5pt} MAE  \cite{mae}    & IN1K & 1600 & 85.80M & - & \graybase{67.51} & \graybase{44.47}  & \graybase{47.35} & \graybase{43.81} & \graybase{83.89} & \graybase{38.84} \\
         \hline
         \textcolor{gray}{\makecell[l]{Full pre-training from IN1K}} & & & & & & & & & & \\
         \hspace{5pt} MAE  \cite{mae}    & IN1K + MSIP & 100 & 85.80M & 85.80M & \graybase{64.90} & \graybase{43.97} & \graybase{45.36}  & \graybase{43.34} & \graybase{83.00} & \graybase{38.41} \\
         \hspace{5pt} \text{MAE + layerwise-decay lr} \cite{clark2020lrdecay} & IN1K + MSIP & 100 & 85.80M & 85.80M & \graybase{67.03} & \graybase{46.44} & \graybase{47.71} & \graybase{\underline{44.39}} & \graybase{\underline{84.03}} & \graybase{\underline{39.11}} \\
         \hline
         \multicolumn{2}{l}{\textcolor{gray}{\makecell[l]{Pre-training with adapter (PAD)}}} & & & & & & & & & \\
         \hspace{5pt} MAE \cite{mae} & IN1K + MSIP & 100 & 87.03M & 1.23M & \baseline{68.41}  & \baseline{\textbf{46.89}} & \baseline{\textbf{48.82}} & \baseline{\textbf{45.18}} & \baseline{\textbf{85.51}} & \baseline{\textbf{40.50}} \\
         \hline
    \end{tabular}
    \vspace{-2mm}
    \caption{\textbf{Comparison with baseline paradigms.} Random RoI cropping is utilized by all paradigms except for \textit{cross-domain fune-tuning}. \textit{Total params} is the total encoder parameter count, while \textit{training params} indicates the number of parameters updated in the encoder during pre-training on MSIP. The top two results are marked in \textbf{bold} and \underline{underlined} format. $\dagger$ denotes task-specific architectures. PAD (marked in \colorbox{baselinecolor}{\rule[-0.2ex]{0pt}{1.5ex}lavender blue}) outperforms other baseline paradigms based on MAE (marked in \colorbox{graycolor}{\rule[-0.2ex]{0pt}{1.5ex}gray}) in all three downstream tasks.}
    \label{tab:comparsion}
    \vspace{-5mm}
\end{table*}

\textbf{Pre-training dataset.}  We construct a \textbf{M}ulti-\textbf{S}cene \textbf{I}nfrared \textbf{P}re-training (MSIP) dataset consisting of 178,756 infrared images, sourced from eight infrared detection and tracking datasets \cite{dronevehicle, scut, jia2021llvip, aerial, ship, security, liu2020lsotb, li2021lasher}. MSIP includes a wide range of scenes such as driving, surveillance, drone, and maritime views, with various objects, including cars, pedestrians, animals, ships, planes, and backgrounds like buildings, forests, rivers, sky, and sea. To reduce redundancy in MSIP, we utilize frame subsampling for image sequences with similar backgrounds. More details about the extraction process are included in \cref{app:pre-training dataset}.

\noindent
\textbf{Pre-training backbone.} The encoder is built upon ViT-Base \cite{vit}, where the original MLP is replaced by PatchAdaptMLP. The decoder comprises eight transformer blocks \cite{vaswani2017transformer} and a linear layer for predicting pixel values. The mask ratio is 75\% and the default middle dimension $r$ in the patchwise-scale adapter is 64.

\noindent
\textbf{Pre-training strategy.} The parameters of the patchwise-scale adapter and the decoder are allowed to be updated, while other parts are frozen. Additionally, in the early stages of pre-training, the parameters of the patchwise-scale module are frozen. This strategy enables the vanilla adapter to learn basic domain-specific feature extraction ability before jointly optimizing both modules, thus preventing training instability. 

\noindent
\textbf{Downstream tasks and datasets.} Infrared object detection on the FLIR \cite{FLIR} dataset and semantic segmentation on the SODA \cite{soda} and MFNet \cite{ha2017mfnet} datasets, are utilized as downstream tasks for evaluating the pre-trained model. It is noteworthy that there is no overlap between these downstream datasets and MSIP. More details are given in \cref{app:downstream dataset}.

\noindent
\textbf{Implementation details.} During pre-training, the model is trained for 100 epochs and the PS module starts updating after 60 epochs. The detection model ViTDet \cite{li2022vitdet} and the segmentation model UperNet \cite{xiao2018upernet} are fine-tuned for 12 and 100 epochs, respectively. More details are in \cref{app:implementation details}.

\subsection{Comparison with baseline methods}

\Cref{tab:comparsion} provides the performance of different paradigms. Despite being pre-trained for 400 epochs, the \textit{full pre-training from scratch} paradigm exhibits the worst performance, which could be attributed to the only exposure to textureless infrared images of the model, resulting in deficient general image feature extraction capability, as shown in the 2nd row of \cref{fig:attention}. In the \textit{cross-domain fine-tuning} paradigm, parameters pre-trained by various image self-supervised and supervised learning methods are employed. It is observed that DeiT \cite{touvron2021deit} outperforms most self-supervised methods in the semantic segmentation task, MILAN \cite{hou2022milan} achieves the highest mIoU on SODA, and MAE \cite{mae} yields competitive results on all three datasets.

It is noteworthy that the \textit{full pre-training from IN1K} paradigm exhibits inferior performance compared to the \textit{cross-domain fine-tuning} paradigm. As demonstrated in the 4th row of \cref{fig:attention}, adjusting all parameters of the model leads to an excessive focus on domain-specific features while forgetting some of its general feature extraction capability. To mitigate this overfitting issue, the layerwise-decay learning rate \cite{clark2020lrdecay} is incorporated. With this approach, the \textit{full pre-training from IN1K} paradigm outperforms the \textit{cross-domain fine-tuning} paradigm on MFNet and FLIR.

\begin{table}[t]
    \centering
    \setlength{\tabcolsep}{0.9mm}{
    \footnotesize
    \begin{tabular}{c|ccc|c|l|ll}
        \hline
         & \multirow{2}{*}{VA} & \multirow{2}{*}{\text{\makecell[c]{PS}}} & \multirow{2}{*}{\text{\makecell[c]{RRC}}} & \multirow{2}{*}{\makecell[c]{Params}} & \multicolumn{1}{c|}{\multirow{2}{*}{SODA}} & \multicolumn{2}{c}{MFNet} \\
          & & & & & & \multicolumn{1}{c}{val} & \multicolumn{1}{c}{test}  \\
        \hline
        (a) & \xmark & \xmark & \xmark & 0.00M & 67.51 & 44.47 & 47.35 \\

        (b) & \cmark & \xmark & \xmark & 1.21M & 67.06\textcolor{gray}{(-0.45)} & 46.11\textcolor{blue}{(+1.64)} & 48.18\textcolor{blue}{(+0.83)} \\
        
        (c) & \cmark & \xmark & \cmark & 1.21M  & 67.80\textcolor{blue}{(+0.29)} & 46.51\textcolor{blue}{(+2.04)} & 48.22\textcolor{blue}{(+0.87)}  \\
        
        (d) & \cmark & \cmark & \xmark & 1.23M  & 67.92\textcolor{blue}{(+0.41)} & 46.11\textcolor{blue}{(+1.64)}  & 48.68\textcolor{blue}{(+1.33)}    \\

        (e) & \cmark & \cmark & \cmark & 1.23M  & \graybase{\textbf{68.41}\textcolor{blue}{(+0.90)}} & \graybase{\textbf{46.89}\textcolor{blue}{(+2.42)}} & \graybase{\textbf{48.82}\textcolor{blue}{(+1.47)}}  \\
        \hline
    \end{tabular}}
    \vspace{-2mm}
    \caption{\textbf{Ablations for components of PAD and random RoI cropping}. VA, PS, and RRC respectively refer to the vanilla adapter, patchwise-scale module, and random RoI cropping. The scale factor in VA is 0.5. Using all components achieves the best performance. The default entry is marked in \colorbox{graycolor}{\rule[0.1pt]{0pt}{0.2pt}gray}.}
    \label{tab:ablation_component}
    \vspace{-2mm}
\end{table}

\begin{table}[t]
    \centering
    \setlength{\tabcolsep}{0.3mm}{
    \footnotesize
    \begin{tabular}{lc|l|ll}
         \hline
        \multirow{2}{*}{Method} & \multirow{2}{*}{RRC} & \multicolumn{1}{c|}{\multirow{2}{*}{SODA}} & \multicolumn{2}{c}{MFNet}\\
         &  &   & \multicolumn{1}{c}{val} & \multicolumn{1}{c}{test} \\
         \hline
         \textcolor{gray}{{\makecell[l]{Pre-training from scratch}}} &  &  &  &  \\
         \hspace{1pt} MAE \cite{mae}  & \xmark & 58.95 & 35.23 &  38.81 \\
         \hspace{1pt} MAE  & \cmark & \graybase{61.46\textcolor{blue}{(+2.51)}} & \graybase{39.58\textcolor{blue}{(+4.35)}} &  \graybase{42.98\textcolor{blue}{(+4.17)}} \\
         \hline
         \textcolor{gray}{{\makecell[l]{Pre-training from IN1K}}} &  &  &  &  \\
         \hspace{1pt} MAE \cite{mae} & \xmark & 63.98 & 39.27 &  43.74 \\
         \hspace{1pt} MAE  & \cmark & \graybase{64.90\textcolor{blue}{(+0.92)}} & \graybase{43.97\textcolor{blue}{(+4.70)}} &  \graybase{45.36\textcolor{blue}{(+1.62)}} \\
         \hspace{1pt} \text{MAE+layerwise-decay lr}   & \xmark & 65.99 & 43.45 & 46.90 \\
         \hspace{1pt} \text{MAE+layerwise-decay lr}  & \cmark & \graybase{67.03\textcolor{blue}{(+1.04)}} & \graybase{46.44\textcolor{blue}{(+2.99)}} &  \graybase{47.71\textcolor{blue}{(+0.81)}} \\
         \hline
    \end{tabular}}
    \vspace{-2mm}
    \caption{\textbf{Ablations for random RoI cropping in baseline paradigms.} Random RoI cropping is beneficial to both \textit{full pre-training from scratch} and \textit{full pre-training from IN1K} paradigms.}
    \label{tab:ablation_rrc}
    \vspace{-5mm}
\end{table}

The proposed PAD paradigm consistently outperforms three baseline paradigms based on MAE across all three datasets. Compared to the \textit{cross-domain fine-tuning}, PAD achieves a 0.9\% increase in mIoU on SODA, a 2.42\% and 1.47\% mIoU boost on the validation and test sets of MNFet, respectively. On FLIR, PAD exhibits improvements in AP, AP50, and AP75 by 1.51\%, 1.73\%, and 2.29\%, respectively. Notably, PAD achieves these results with only 1.23M trainable parameters in the encoder, surpassing the \textit{full pre-training from IN1K} paradigm that tunes the entire 85.8M parameters of the encoder. MAE-based PAD also outperforms networks specifically designed for downstream tasks by a large margin. PAD achieves a 4.61\% higher mIoU than EC-CNN \cite{soda} on SODA and a 7.94\% higher AP50 than SSTN101 \cite{munir2021sstn} on FLIR. For MFNet, originally designed for paired infrared and RGB semantic segmentation, specific results using only infrared images are unavailable.

\subsection{Ablation studies}

All ablation experiments are conducted on the SODA and MFNet datasets based on ViT-Base and MAE, and the models are pre-trained for 100 epochs. 

\noindent \textbf{Components in PAD.} 
The effectiveness of the proposed patchwise-scale module and random RoI cropping is verified in \cref{tab:ablation_component}. Compared to \textit{cross-domain fine-tuning}, PAD with the vanilla adapter shows notable improvements on MFNet but a performance decrease on SODA. Adding either the patchwise-scale module or the random RoI cropping can enhance the performance, with the best results obtained when both of them are employed.

\noindent \textbf{Random RoI cropping in baseline paradigms.} Random RoI cropping is also beneficial to baseline paradigms. As shown in \cref{tab:ablation_rrc}, with random RoI cropping, the performance of \textit{full pre-training from scratch} and \textit{full pre-training from IN1K} both improve significantly, demonstrating the strong generalization ability of this method. 

\begin{table}[t]
    \centering
    \setlength{\tabcolsep}{1.5mm}{
    \footnotesize
    \begin{tabular}{c|cccc|c|cc}
        \hline
         & \multirow{2}{*}{VA} & \multirow{2}{*}{\text{\makecell[c]{VA \\ pre-train}}} & \multirow{2}{*}{\text{\makecell[c]{PS}}} & \multirow{2}{*}{\text{\makecell[c]{PS \\ pre-train}}} & \multirow{2}{*}{SODA} & \multicolumn{2}{c}{MFNet}\\
         & & & & & & val & test \\
         \hline
         (a) & \xmark & \xmark & \xmark & \xmark & 67.51 & 44.47 & 47.35 \\
         \hline
         (b) & \cmark & \xmark & \xmark & \xmark & 67.18 & 45.00 & 47.29 \\
         (c) & \cmark & \cmark & \xmark & \xmark & 67.80 & 46.51 & 48.22 \\
         \hline
         (d) & \cmark & \xmark & \cmark & \xmark & 67.90 & 45.30  & 47.28 \\
         (e) & \cmark & \cmark & \cmark & \xmark & 68.10 & 46.33 & 48.63 \\
         (f) & \cmark & \cmark & \cmark & \cmark & \graybase{\textbf{68.41}} & \graybase{\textbf{46.89}}  & \graybase{\textbf{48.82}} \\
         \hline 
    \end{tabular}}
    \vspace{-2mm}
    \caption{\textbf{Ablations for pre-training of the patchwise-scale adapter.} Random RoI cropping is utilized. The entry (a) represents the \textit{cross-domain fine-tuning} paradigm. Pre-training is essential for the vanilla adapter and the patchwise-scale module.}
    \label{tab:ablation_pretrain}
    \vspace{-2mm}
\end{table}

\begin{table}[t]
    \centering
    \footnotesize
    \begin{tabular}{cc|c|cc}
        \hline
        \multirow{2}{*}{Input} & \multirow{2}{*}{Params} & \multirow{2}{*}{SODA} & \multicolumn{2}{c}{MFNet}\\
        &  &  & val & test \\
        \hline
        concat & 1.23M & \graybase{\textbf{68.41}} & \graybase{\textbf{46.89}} & \graybase{\textbf{48.82}} \\
        sum  & 1.22M & 68.02 & 46.24 & 48.60 \\
        adapt only & 1.22M & 67.91 & 46.17 & 48.38 \\
        mlp only & 1.22M & 67.96 & 46.02 & 48.41 \\
        \hline
    \end{tabular}
    \vspace{-2mm}
    \caption{\textbf{Ablations for the input to the patchwise-scale module.} \textit{Adapt only} and \textit{mlp only} refer to using only the adapter output or the original MLP output as the input. Concatenating the output of both branches performs best.}
    \label{tab:ablation_patchwise}
    \vspace{-5mm}
\end{table}

\noindent \textbf{Pre-training Necessity for the patchwise-scale adapter.} Firstly, pre-training is essential for the vanilla adapter. Compared to \textit{cross-domain fine-tuning} (\cref{tab:ablation_pretrain} (a)), adding the vanilla adapter directly to downstream tasks (\cref{tab:ablation_pretrain} (b)) without pre-training does not enhance the performance and may even result in worse results. On the contrary, pre-training the vanilla adapter on MSIP and then applying it to downstream tasks (\cref{tab:ablation_pretrain} (c)) achieves effective enhancement in performance, implying that pre-training enables the adapter to acquire domain-specific features extraction capability. Similarly, for the patchwise-scale adapter, pre-training only the vanilla adapter (\cref{tab:ablation_pretrain} (e)) surpasses its counterpart without pre-training (\cref{tab:ablation_pretrain} (d)). Secondly,  pre-training the patchwise-scale module is also vital. Although directly applying the patchwise-scale module for downstream tasks without pre-training (\cref{tab:ablation_pretrain} (e)) outperforms models without this module (\cref{tab:ablation_pretrain} (c)), pre-training (\cref{tab:ablation_pretrain} (f)) further improves the performance, indicating that the variability in scale factors learned during pre-training is beneficial for downstream tasks. Moreover, using the patchwise-scale adapter without pre-training (\cref{tab:ablation_pretrain} (d)), also outperforms the vanilla adapter (\cref{tab:ablation_pretrain} (b)) and the \textit{cross-domain fine-tuning} paradigm (\cref{tab:ablation_pretrain} (a)), demonstrating the importance of the patchwise-scale module.

\begin{table}[t]
    \centering
    \setlength{\tabcolsep}{1.5mm}{
    \footnotesize
    \begin{tabular}{c| c| c |c|cc}
    \hline
         & \multirow{2}{*}{Scale factor} & \multirow{2}{*}{Init value} & \multicolumn{1}{c|}{\multirow{2}{*}{SODA}} & \multicolumn{2}{c}{MFNet}  \\
         & &  &   & \multicolumn{1}{c}{val} & \multicolumn{1}{c}{test} \\
         \hline
         \multirow{5}{*}{(a)} & \multirow{5}{*}{Layerwise frozen} & 0.1  &  68.09  &  46.16   &   48.37 \\
         & &  0.5  &  67.80  & 46.51  &  48.22 \\
         & &  1.0  &  67.76  & 46.27  &  48.49 \\
         & & exp-decay & 67.46  & 46.20  & 48.15  \\
         & & even-decay & 67.83  & 46.02  &  48.18 \\
         \hline
         \multirow{3}{*}{(b)} & \multirow{3}{*}{Layerwise learnable}  &  0.1   & 68.03 &  46.38  &   48.60 \\
         & &  0.5   &  67.70   &  45.74  &  48.63  \\
         & &  1.0   &  67.76   &  45.18  &  48.23  \\
         \hline
         (c) & Patchwise  &  -  &  \graybase{\textbf{68.41}}  &  \graybase{\textbf{46.89}}  & \graybase{\textbf{48.82}} \\
         \hline
    \end{tabular}}
    \vspace{-2mm}
    \caption{\textbf{Ablations for scale factors in the adapter.} \textit{Layerwise frozen} indicates that the scale is layerwise and frozen, while \textit{layerwise learnable} indicates that the scale is layerwise and trainable. \textit{Exp-decay} and \textit{even-decay} represent exponential decay and uniform decay of the scale from deep to shallow layers, respectively. The decay rate is 0.8 and the scale of the last layer is 1.0. The patchwise-scale adapter performs best.}
    \vspace{-3mm}
    \label{tab:ablation_scale}
\end{table}

\begin{table}[t]
    \centering
    \setlength{\tabcolsep}{1.2mm}{
    \footnotesize
    \begin{tabular}{cc|l|ll}
        \hline
        \multirow{2}{*}{Dim} & \multirow{2}{*}{Params} & \multicolumn{1}{c|}{\multirow{2}{*}{SODA}} & \multicolumn{2}{c}{MFNet}\\
         &  &   & \multicolumn{1}{c}{val} & \multicolumn{1}{c}{test} \\
        \hline
        N/A & 0.00M & 67.51 & 44.47 & 47.35 \\
        1 & 0.06M & 67.89\textcolor{blue}{(+0.38)} & 45.63\textcolor{blue}{(+1.16)} & 48.12\textcolor{blue}{(+0.77)} \\
        8 & 0.20M & 67.92\textcolor{blue}{(+0.41)} & 45.99\textcolor{blue}{(+1.52)} & 48.51\textcolor{blue}{(+1.16)} \\
        32 & 0.63M & 68.19\textcolor{blue}{(+0.68)} & 46.80\textcolor{blue}{(+2.33)} & 48.27\textcolor{blue}{(+0.92)}  \\
        64 & 1.23M & \graybase{\textbf{68.41}\textcolor{blue}{(+0.90)}} & \graybase{\textbf{46.89}\textcolor{blue}{(+2.42)}} & \graybase{\textbf{48.82}\textcolor{blue}{(+1.47)}}  \\
        \hline
    \end{tabular}}
    \vspace{-2mm}
    \caption{\textbf{Ablations for middle dimensions of the patchwise-scale adapter.} N/A refers to the \textit{cross-domain fine-tuning} paradigm. The adapter with higher dimensions has more trainable parameters and performs better.}
    \label{tab:ablation_dim}
    \vspace{-5mm}
\end{table}

\noindent \textbf{Input to the patchwise-scale module.} As indicated in \cref{tab:ablation_patchwise}, using the output from either branch as input or simply adding the two branches together is not as effective as concatenating the two branches, implying that considering features from both branches and assessing the relative importance of domain-specific features is more advantageous.

\begin{figure*}[t]
  \vspace{-2mm}
  \centering
  \includegraphics[width=1.0\linewidth]{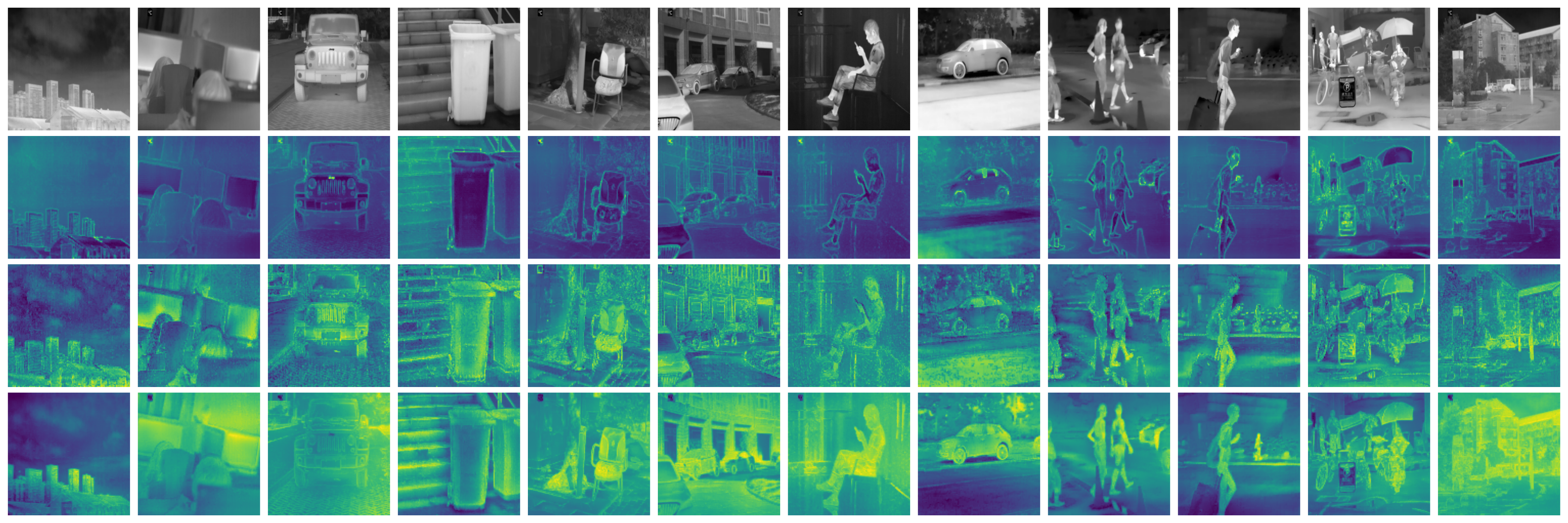}
  \vspace{-6mm}
  \caption{\textbf{Visualization of original images and normalized scale factor maps from the patchwise-scale modules in the pre-trained encoder.} The first row represents the original images from SODA. The 2nd to 4th rows depict the normalized scale factor maps for the 2nd, 9th, and 12th layers in the pre-trained model. The patchwise-scale module generalizes well to previously unseen images.}
  \label{fig:scalar}
  \vspace{-2mm}
\end{figure*}

\begin{table*}[t]
    \centering
    \footnotesize
    \begin{tabular}{cc|l|ll|lll}
        \hline
        \multirow{2}{*}{Method} & \multirow{2}{*}{Paradigm} & \multicolumn{1}{c|}{\multirow{2}{*}{SODA}} & \multicolumn{2}{c|}{MFNet} & \multicolumn{3}{c}{FLIR}\\
         &  & & \multicolumn{1}{c}{val} & \multicolumn{1}{c|}{test} & \multicolumn{1}{c}{AP} & \multicolumn{1}{c}{AP50} & \multicolumn{1}{c}{AP75}\\
        \hline
        \multirow{3}{*}{MILAN \cite{hou2022milan}} & Cross-domain fine-tuning  & 68.93 & 45.08 & 47.11 & 42.29 & 81.99 & 36.46 \\
         & Pre-training from IN1K & 69.00\textcolor{blue}{(+0.07)}  & 46.27\textcolor{blue}{(+1.19)}  &  \textbf{48.53}\textcolor{blue}{(+1.42)} & 42.48\textcolor{blue}{(+0.19)}  &  82.76\textcolor{blue}{(+0.77)}  & 36.51\textcolor{blue}{(+0.05)} \\
         & Pre-training with adapter & \textbf{69.16}\textcolor{blue}{(+0.23)}  & \textbf{48.02}\textcolor{blue}{(+2.94)} & 47.83\textcolor{blue}{(+0.72)} & \textbf{43.23}\textcolor{blue}{(+0.94)} & \textbf{83.59}\textcolor{blue}{(+1.60)} & \textbf{37.70}\textcolor{blue}{(+1.24)} \\
        \hline
    \end{tabular}
    \vspace{-2mm}
    \caption{\textbf{The performance of PAD based on MILAN.} Random RoI cropping and layerwise-decay learning rate are utilized in \textit{Pre-training from IN1K}. The PAD pre-training paradigm is beneficial for other self-supervised learning methods.}
    \label{tab:other method}
    \vspace{-5mm}
\end{table*}

\noindent \textbf{Scale factor.} \Cref{tab:ablation_scale} investigates the impact of different scale factor configurations. When employing an identical and fixed scale factor for each layer (the first three entries of \cref{tab:ablation_scale} (a)), smaller scales perform better on SODA, while larger values are more effective on MFNet. This observation may be attributed to the inclusion of many indoor-scene infrared images in SODA, which are less common in the pre-training dataset, resulting in a larger domain gap and a reduced need for pre-trained domain-specific features. Setting smaller scales for shallow layers and larger scales for deep layers (the last two entries of \cref{tab:ablation_scale} (a)) fails to surpass the performance of using the same scale factor across all layers. When introducing learnable layerwise scale factors (\cref{tab:ablation_scale} (b)), no evident advantage is observed compared to the fixed scale, confirming the discussion in \cref{subsec:limitations} that scale factors learned during pre-training may not be suitable for downstream tasks. In contrast, the patchwise scale factor demonstrates better performance, indicating that the variability in scale factors of different patches learned during pre-training is advantageous for downstream tasks.

\noindent \textbf{Middle dimension.} The middle dimension of the adapter determines its overall parameter count \cite{chen2022adaptformer}. \Cref{tab:ablation_dim} provides the performance of the patchwise-scale adapter with varying middle dimensions. Notably, when the dimension is 1, PAD already outperforms \textit{cross-domain fine-tuning}, with only 0.06M extra parameters. The Performance continues to improve with the increasing dimension.

\subsection{Other self-supervised methods}
PAD is also applicable to other SSL methods. As indicated in \cref{tab:other method}, based on MILAN \cite{hou2022milan}, which leverages the image encoder of CLIP \cite{radford2021clip} to generate semantic image features as supervisory signals, PAD outperforms the \textit{cross-domain fine-tuning} and \textit{pre-training from IN1K} paradigms overall, demonstrating its strong applicability.

\subsection{Visualization}
The outputs of the patchwise-scale module in the pre-trained encoder is visualized in \cref{fig:scalar}. The scale factor maps of different layers exhibit distinct patterns. Shallow layers, such as the 2nd layer (the 2nd row of \cref{fig:scalar}), tend to focus on local information like edges in the image. As the depth increases, the module shifts its focus to global information, and the scale map gradually reveals some semantic patterns, with different patches of the same object or similar objects having close scale values, as observed in the 9th and 12th layers (the last two rows of \cref{fig:scalar}). It is noteworthy that the images in \cref{fig:scalar} are from SODA, which are not visible to the pre-trained model. Nevertheless, the patchwise-scale module still generates meaningful scale factor maps, indicating its strong generalization capability. The histograms of the scale factor in different tasks are given in \cref{app:visualization}.

\section{Conclusion}
\label{sec:conclusion}

In this paper, we investigate several key challenges in SSL with infrared images. We construct an infrared pre-training dataset MSIP and introduce an image preprocessing method tailored to non-iconic images. By making a comparison of three baseline pre-training paradigms, the PAD paradigm and the patchwise-scale adapter are further proposed. Extensive experiments with various SSL methods reveal that pre-training with the patchwise-scale adapter, involving only 1.23M pre-trainable parameters, surpasses continual pre-training of the entire model, demonstrating the effectiveness and generalization ability of this paradigm. It is believed that PAD can be extended beyond the infrared imagery and applied to other data-scarce domains. We expect that our work can inspire further research in this field.

\noindent \textbf{Limitations and future work.} Due to the dynamic nature, it is nontrivial to fuse the proposed patchwise-scale adapter into the original network after pre-training, thus introducing a small number of additional parameters during inference and downstream fine-tuning. In the future, we will investigate the solution for parameter fusion under the dynamic constraint in our PAD paradigm. 

{
    \small
    \bibliographystyle{ieeenat_fullname}
    \bibliography{main}
    
}
\clearpage
\setcounter{page}{1}
\maketitlesupplementary
\appendix

\section{Influence of the scale factor}
\label{app:scale factor}
The scale factor in the adapter plays two roles. Firstly, the scale factor affects the learning speed of the parameters in the adapter. Let's consider the updating process of the weight in the up-projection layer as an example. For simplicity, the bias term is omitted. Assuming it is currently in the $i$-th step of parameter updating, let $x$, $W_{\text{up}}^i\in \mathbb{R}^{d\times r}$, $y^i\in \mathbb{R}^r$ denote the input, weight, and output of the up-projection layer in this step, respectively. The $s$, $\eta^i$, and $L$ represent the scale factor, the learning rate in this step, and the loss function, respectively. Thus, the output of the up-projection layer can be formulated as follows:
\begin{equation}
    y^i=s \cdot W_{\text{up}}^i x.
\end{equation}
The gradient of the weight $W_{\text{up}}^i$ in the step is
\begin{equation}
    \frac{\partial L}{\partial W_{\text{up}}^i}=\frac{\partial L}{\partial y^i} \frac{\partial y^i}{\partial W_{\text{up}}^i} = s \cdot \frac{\partial L}{\partial y^i} \frac{\partial (W_{\text{up}}^ix)}{\partial W_{\text{up}}^i},
\end{equation}
and the updating amount of $W_{\text{up}}^i$ is
\begin{equation}
    \Delta W_{\text{up}}^i=-\eta^i \cdot \frac{\partial L}{\partial W_{\text{up}}^i}=-s\eta^i \cdot \frac{\partial L}{\partial y^i} \frac{\partial (W_{\text{up}}^ix)}{\partial W_{\text{up}}^i}.
    \label{eq:update_w}
\end{equation}
In the $i+1$-th step, the value of $W_{\text{up}}^{i+1}$ becomes
\begin{equation}
    W_{\text{up}}^{i+1}=W_{\text{up}}^i+\Delta W_{\text{up}}^i,
\end{equation}
and the output $y^{i+1}$ is
\begin{equation}
\begin{split}
    y^{i+1}&= s \cdot W_{\text{up}}^{i+1}x=s \cdot W_{\text{up}}^ix+s \cdot \Delta W_{\text{up}}^i \\
    &=y^i -\eta^i s^2 \left( \frac{\partial L}{\partial y^i} \frac{\partial (W_{\text{up}}^ix)}{\partial W_{\text{up}}^i} \right) x.
    \label{eq:update_y}
\end{split}
\end{equation}
When there is no scale factor, \ie $s$ equals 1, the \cref{eq:update_w} can be rewritten as follows:
\begin{equation}
    \Delta W_{\text{up}}^i=-\eta^i \cdot \frac{\partial L}{\partial W_{\text{up}}^i}=-\eta^i \cdot \frac{\partial L}{\partial y^i} \frac{\partial (W_{\text{up}}^ix)}{\partial W_{\text{up}}^i},
    \label{eq:update_w1}
\end{equation}
and the \cref{eq:update_y} becomes
\begin{equation}
    y^{i+1}=y^i -\eta^i \left( \frac{\partial L}{\partial y^i} \frac{\partial (W_{\text{up}}^ix)}{\partial W_{\text{up}}^i} \right) x.
    \label{eq:update_y1}
\end{equation}

\begin{figure}[t]
  \vspace{-2mm}
  \centering
  \includegraphics[width=1.0\linewidth]{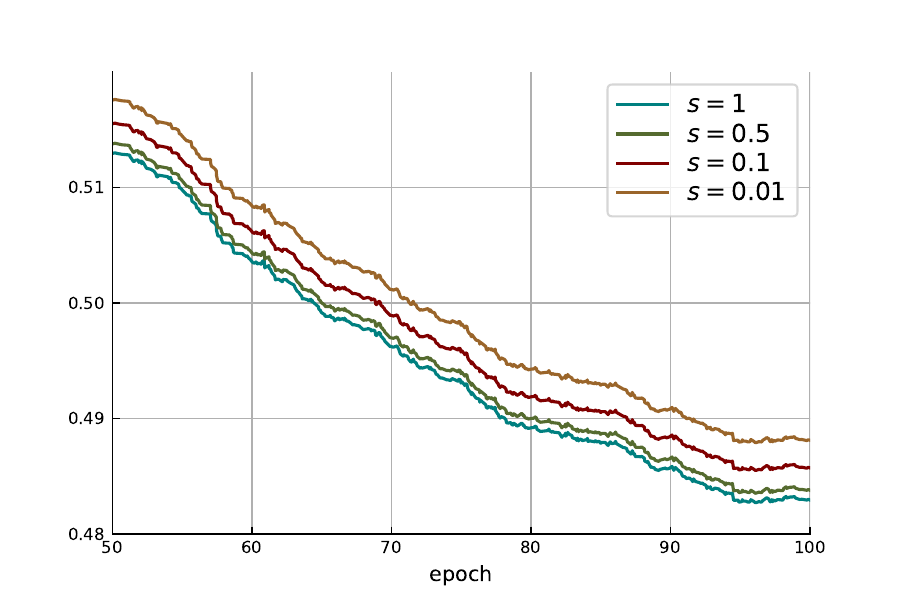}
  \vspace{-6mm}
  \caption{\textbf{Smoothed pre-training loss for different scale factors.} The pre-training loss with a larger $s$ decreases more rapidly.} 
  \label{fig:loss}
  \vspace{-3mm}
\end{figure}

\begin{figure}[t]
  \centering
  \includegraphics[width=1.0\linewidth]{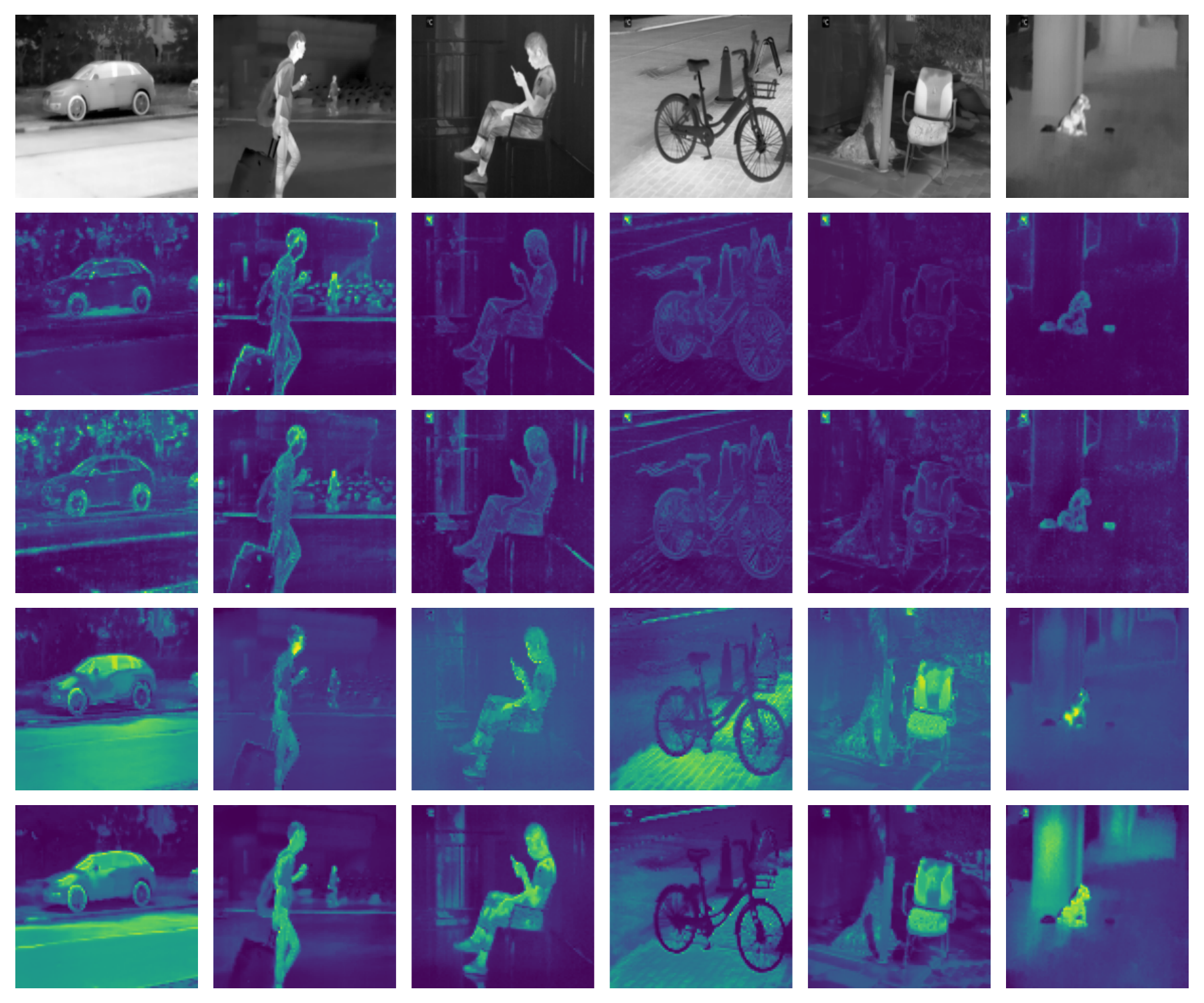}
  \vspace{-6mm}
  \caption{\textbf{Visualizations of original images and attention maps extracted from the last self-attention layer of the pre-trained model.} The model is pre-trained with the vanilla adapter using scale factors of 0.5, while different scale factors are used during inference. The 2nd row to the 5th row corresponds to scale factors of 0.01, 0.5, 5, and 10, respectively.} 
  \label{fig:attention_scale}
  \vspace{-6mm}
\end{figure}

Comparing \cref{eq:update_w} and \cref{eq:update_w1}, as well as \cref{eq:update_y} and \cref{eq:update_y1}, it is observed that the inclusion of the scale factor accelerates the update rate of the parameters in the adapter by a factor of $s$ and speeds up the convergence of the adapter's output to the target value by a factor of $s^2$. As depicted in \cref{fig:loss}, it is evident that the pre-training loss decreases more rapidly with larger scale factors. This observation holds for various scale values, as the initialization strategy of the adapter remains consistent across different scale values.

\begin{table*}[t]
    \centering
    \footnotesize
    \begin{tabular}{l | c c c | c c c c}
        \hline
        Dataset & \#Images & \makecell[c]{Average width \\ of images} & \makecell[c]{Average height \\ of images} & \#RoIs & \makecell[c]{Average width \\ of RoIs} & \makecell[c]{Average height \\ of RoIs} & Average ratio \\
        \hline
        DroneVehicle \cite{dronevehicle}         & 17,990  & 640  & 512  &  57  & 176  & 183  & 0.15 \\
        SCUT FIR Pedestrian \cite{scut} & 21,049  & 720  & 576  &  20  & 393  & 311  & 0.38 \\
        LLVIP   \cite{jia2021llvip}             & 12,025  & 1280 & 1024 &  143 & 217  & 191  & 0.06 \\
        Infrared Aerial Photography \cite{aerial} &  5,523  & 627  & 502  &  59  &  212  & 158  &  0.16 \\
        Infrared Ship   \cite{ship}     & 8,402   & 772  & 592  &  25  & 368  & 211  & 0.34 \\
        Infrared Security  \cite{security}  & 8,999  & 424  & 323  &  14  &  216  & 142  & 0.31  \\
        LSOTB-TIR     \cite{liu2020lsotb}       & 52,925  & 913  & 618  & 27  & 289  & 219  & 0.26 \\
        LasHeR    \cite{li2021lasher}           & 51,843  & 873  & 552  & 18  & 324  & 237  & 0.26 \\
        \hline
    \end{tabular}
    \vspace{-2mm}
    \caption{\textbf{Image and RoI statistics for sub-datasets of the pre-training dataset MSIP.} \#Images represents the number of images in each sub-dataset and \#RoIs indicates the average number of candidate RoIs generated by the selective search in each image. The average ratio signifies the average proportion of the candidate RoI area to the entire image area.}
    \label{tab:dataset}
    \vspace{-5mm}
\end{table*}

In addition, the scale factor also controls the importance of domain-specific features. For the same pre-trained model, significant changes in the attention maps are observed when using different scale factors during inference. In \cref{fig:attention_scale}, the scale factors in the model are maintained at 0.5 during pre-training, while various scale values are utilized during inference. It is observed that with an increase in the scale value, the attention of the pre-trained model shifts towards the brighter regions in the image, leading to a reduced emphasis on edge information. According to this, it can be inferred that the adapter is primarily responsible for extracting domain-specific features. As the scale factor increases, the influence of domain-specific features becomes more pronounced in the model.

\section{Datasets}
\label{app:datasets}
\subsection{Pre-training datasets}
\label{app:pre-training dataset} 

Our pre-training dataset MSIP consists of 178,756 images extracted from the following datasets:

\noindent \textbf{DroneVehicle \cite{dronevehicle}.} The DroneVehicle dataset is a vehicle detection dataset captured from the aerial perspective of drones. It comprises 28,439 image pairs, with each pair containing both infrared and RGB images. The dataset encompasses various scenes, including urban roads, residential areas, and parking lots, with images captured during both daytime and nighttime. For our pre-training dataset, all 17,990 infrared training images are extracted.

\noindent \textbf{SCUT FIR Pedestrian \cite{scut}.} This dataset is a sizable dataset specifically designed for far-infrared pedestrian detection. It consists of a collection of image sequences spanning 11 hours, resulting in a total of 211,011 annotated images. These images represent a variety of urban scenes, including city centers, suburban areas, highways, and campus environments. To reduce redundancy arising from the high similarity between adjacent images, we employ a subsampling strategy, selecting one image out of every ten, which yields a refined sub-dataset consisting of 21,049 images.

\noindent \textbf{LLVIP \cite{jia2021llvip}.} The LLVIP dataset is an RGBT pedestrian detection dataset captured from road monitoring perspectives. It contains a total of 15,488 image pairs. The entire collection of infrared images from the training set is utilized in MSIP, totaling 12,025 images.

\noindent \textbf{Infrared Aerial Photography \cite{aerial}.} This dataset comprises 11,045 infrared images captured from an unmanned aerial vehicle (UAV) perspective. These images cover a variety of scenes, such as roads, residential areas, and beaches. Due to the high similarity between images, one image for every two is extracted, resulting in a total of 5,523 images.

\noindent \textbf{Infrared Ship \cite{ship}.} This dataset consists of 9,402 infrared images primarily curated for ship detection. It includes various types of ships in diverse maritime environments, such as seas, ports, and coastal areas. All 8,402 training images are utilized in MSIP.

\noindent \textbf{Infrared Security \cite{security}.} This dataset is comprised of 8,999 images focused on infrared human and vehicle detection from a top-down surveillance perspective, representing real-world security scenarios. All 8,999 images are used in our pre-training dataset. 

\noindent \textbf{LSOTB-TIR \cite{liu2020lsotb}.} The LSOTB-TIR dataset is a large-scale collection of infrared target-tracking data. It encompasses 1,400 sequences of infrared images, comprising over 600,000 individual frames. For MSIP, we concentrate on 1,280 training sequences and extract one frame out of every ten, resulting in a total of 52,925 images.

\noindent \textbf{LasHeR \cite{li2021lasher}.} The LasHeR dataset is a large-scale RGBT tracking dataset, consisting of 1,224 pairs of infrared and RGB image sequences. It includes over 730,000 image pairs covering diverse scenes and object categories. We utilize one image out of every ten from its 979 infrared training sequences, obtaining 51,843 images.

\begin{figure*}[p]
  \centering
  \vspace{1mm}
  \includegraphics[width=1.0\linewidth]{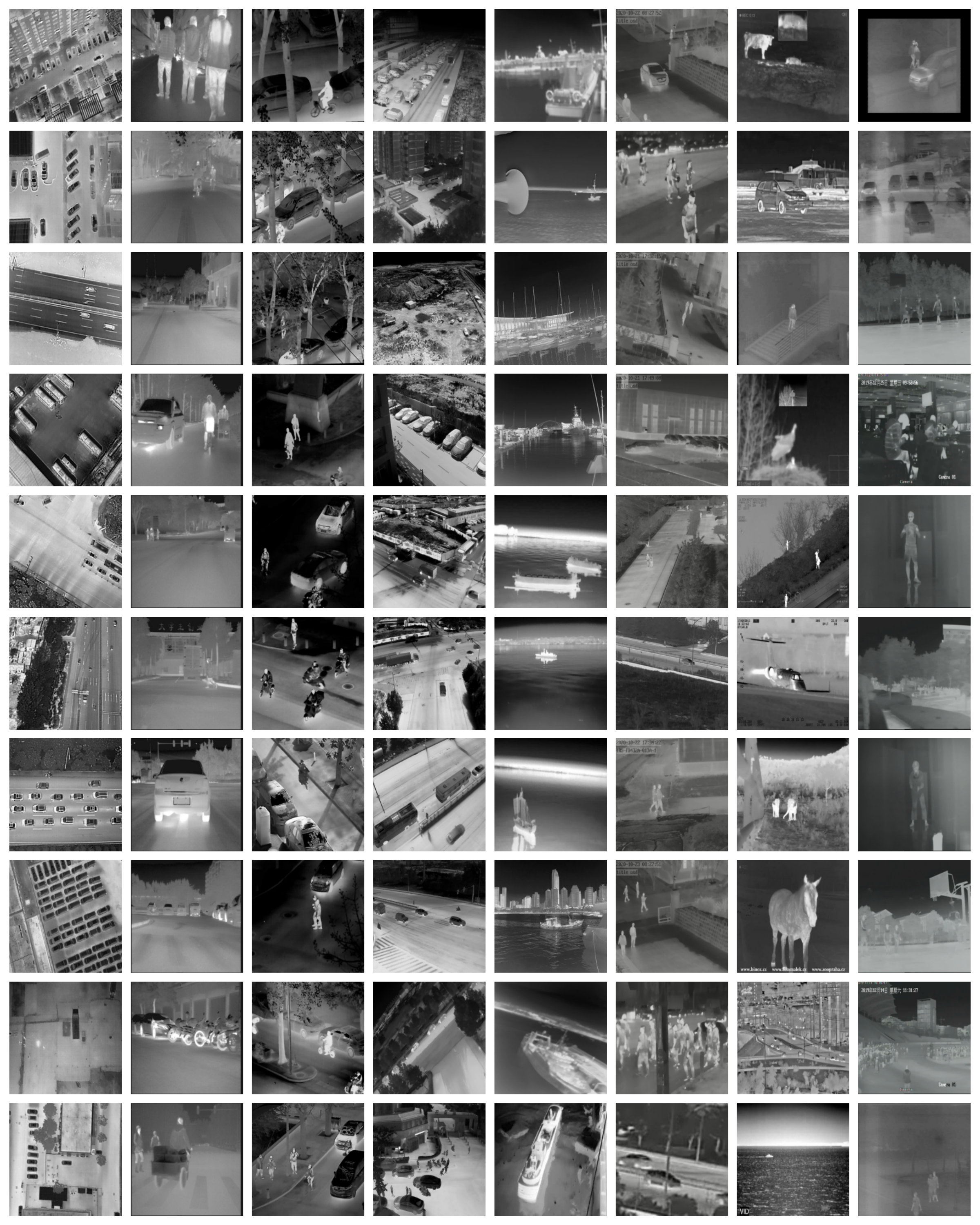}
  \vspace{-6mm}
  \caption{\textbf{Example images from the pre-training dataset MSIP.} Each column represents a sub-dataset, arranged from left to right according to the order in \cref{tab:dataset}. All images are resized to the same size.} 
  \label{fig:sample_img}
  \vspace{-6mm}
\end{figure*}

The statistical information of these extracted sub-datasets is presented in \cref{tab:dataset}, and the corresponding example images are illustrated in \cref{fig:sample_img}.

\subsection{Downstream datasets}
\label{app:downstream dataset}
\noindent \textbf{FLIR \cite{FLIR}.} This dataset is designed for infrared object detection in automotive driving scenarios and comprises 8,862 training images and 1,366 test images. It encompasses three primary object categories: people, car, and bicycle.

\noindent \textbf{SODA \cite{soda}.} This dataset covers various indoor and outdoor scenes for infrared semantic segmentation. It includes 1,168 training images and 1,000 test images, encompassing 20 different semantic categories such as road, building, car, chair, lamp, table, monitor, and others. 

\noindent \textbf{MFNet \cite{ha2017mfnet}.} The MFNet dataset, on the other hand, is focused on RGBT semantic segmentation in automotive driving scenarios. It contains a total of 1,569 image pairs, comprising both infrared and RGB images. In this work, we only use the infrared images, which include 784 training images, 392 validation images, and 393 test images, containing 8 semantic categories like car, person, bike, curve, and others.

\section{Implementation details}
\label{app:implementation details}

\subsection{Random RoI cropping}
\label{app:rrc}

\begin{algorithm}[t] 
	\renewcommand{\algorithmicrequire}{\textbf{Input:}}
	\renewcommand{\algorithmicensure}{\textbf{Output:}}
	\caption{Random RoI Cropping} 
	\label{alg:rrc}
	\begin{algorithmic}[1] 
		\Require 
		$I$ represents a pre-training image. $R$ denotes the set of RoIs in image $I$ generated by selective search \cite{uijlings2013selectivesearch} and $N$ is the number of RoIs in $R$.
        $M$ is the maximum number of iterations and 
        $ l_{\text{min}} $ is the minimum desired width and height. $LargeRoI$ indicates
        whether to select RoIs with large areas. $A$ denotes the area of $I$.
		\Ensure 
		The coordinates of the cropping box $x', y', w', h'$.
        \State $i\gets 0$
        \While{$i < M$}
        \State $j \gets$ Uniformly sample from $[1, N]$;
        \State $x, y, w, h \gets$ Select the $j$-th RoI from $R$;
        \State $x_c \gets x + 0.5w$;
        \State $y_c \gets y + 0.5h$;
        \State $x_c' \gets $ Uniformly sample from $[x_c - 0.4w, x_c + 0.4w]$;
        \State $y_c' \gets $ Uniformly sample from $[y_c - 0.4h, y_c + 0.4h]$;
		\State $w' \gets $ Uniformly sample from $[0.7w, 1.5w]$;
        \State $h' \gets $ Uniformly sample from $[0.7h, 1.5h]$;
        \State $x' \gets x_c' - 0.5w'$;
        \State $y' \gets y_c' - 0.5h'$;
        \State Clamp the region specified by $x', y', w', h'$ to ensure that the box remains entirely within the image;
        \If{$w' < l_{\text{min}} $ or $h' <  l_{\text{min}} $}
            \State Continue;
        \EndIf
        \State $ratio \gets w'\cdot h' / A$
        \If{$LargeRoI$ and $ratio < 0.15$}
            \State Continue;
        \EndIf
        \State Return $x', y', w', h'$;
        \EndWhile
        \State Return RandomResizedCrop($I$).
	\end{algorithmic} 
        
\end{algorithm}
\vspace{-2mm}
As depicted in \cref{alg:rrc}, the selective search \cite{uijlings2013selectivesearch} method is firstly employed to generate multiple RoIs for each image before pre-training. The statistical information of these generated RoIs is provided in \cref{tab:dataset}. During each epoch of pre-training, the random RoI cropping operation randomly selects one of these candidate RoIs. Subsequently, it applies random translations within a specified range to the center of the RoI and scales its width and height within certain bounds. If the side length of the scaled RoI becomes too small or its area relative to the original images diminishes significantly, a new RoI is chosen, and the aforementioned steps are repeated. If the maximum number of iterations is reached without obtaining satisfactory RoIs, random resized cropping is then applied to the input image.

In our experiments, the maximum number of iterations $M$ and the minimum desired side length $l_\text{min}$ are set to 20 and 60, respectively. Furthermore, the probability of $LargeRoI$ being true is set to 15\% to ensure that a certain proportion of high-resolution images are used as input during pre-training.

\begin{table}[t]
    \centering
    \footnotesize
    \begin{tabular}{l | c}
        \hline
        Hyperparameters  &  Value \\
        \hline
        Input resolution  &  224 $\times$  224 \\
        Optimizer  &  AdamW \cite{loshchilov2018adamw} \\
        Base learning rate  & 1e-4  \\
        Weight decay  & 0.05  \\
        Optimizer momentum  &  $\beta_1, \beta_2=$ 0.9, 0.95 \\
        Batch size  &  4096 \\
        Learning rate schedule  &  Cosine decay \cite{loshchilov2016cosinedecay}  \\
        Warmup epochs \cite{goyal2017warmepoch}  &  30  \\
        Training epochs &  100 \\
        Augmentation  &  Random RoI cropping \\
        \hline 
    \end{tabular}
    \vspace{-2mm}
    \caption{\textbf{Infrared pre-training setting.}}
    \label{tab:setting_pretrain}
    \vspace{-2mm}
\end{table}

\begin{table}[t]
    \centering
    \footnotesize
    \begin{tabular}{l | c}
        \hline
        Hyperparameters  &  Value \\
        \hline
        Input resolution  &  1024 $\times$  1024 \\
        Optimizer  &  AdamW \cite{loshchilov2018adamw} \\
        Base learning rate  & 1e-4  \\
        Weight decay  & 0.1  \\
        Optimizer momentum  &  $\beta_1, \beta_2=$ 0.9, 0.999 \\
        Batch size  &  2 \\
        Layerwise-decay learning rate \cite{clark2020lrdecay}  & 0.7 \\
        Learning rate schedule  &  Step decay \\
        Warmup steps  &  250  \\
        Fine-tuning epochs &  12 \\
        \hline 
    \end{tabular}
    \vspace{-2mm}
    \caption{\textbf{Infrared object detection setting.}}
    \label{tab:setting_det}
    \vspace{-2mm}
\end{table}

\begin{table}[t]
    \centering
    \footnotesize
    \begin{tabular}{l | c  c}
         \hline
         Hyperparameters  &  SODA  & MFNet \\
         \hline
         Input resolution  &  \multicolumn{2}{c}{512 $\times$  512} \\
         Peak learning rate  &  \multicolumn{2}{c}{1e-4} \\
         Fine-tuning epochs &  \multicolumn{2}{c}{100} \\
         Fine-tuning steps  &  14400  &  9800  \\
         Batch size  &  \multicolumn{2}{c}{8}  \\
         Optimizer  &  \multicolumn{2}{c}{AdamW}  \\
         Weight decay  & \multicolumn{2}{c}{0.05}  \\
         Optimizer momentum  &  \multicolumn{2}{c}{$\beta_1, \beta_2=$ 0.9, 0.999} \\
         Layerwise-decay learning rate  & \multicolumn{2}{c}{0.65} \\
         Learning rate schedule  & \multicolumn{2}{c}{Linear decay} \\
         Minimal learning rate & \multicolumn{2}{c}{0} \\
         Warmup steps  & 1500 &  1000 \\
         \hline
    \end{tabular}
    \vspace{-2mm}
    \caption{\textbf{Infrared semantic segmentation setting.}}
    \label{tab:setting_seg}
    \vspace{-5mm}
\end{table}

\begin{figure*}[t]
  \centering
  \vspace{-2mm}
  \includegraphics[width=1.0\linewidth]{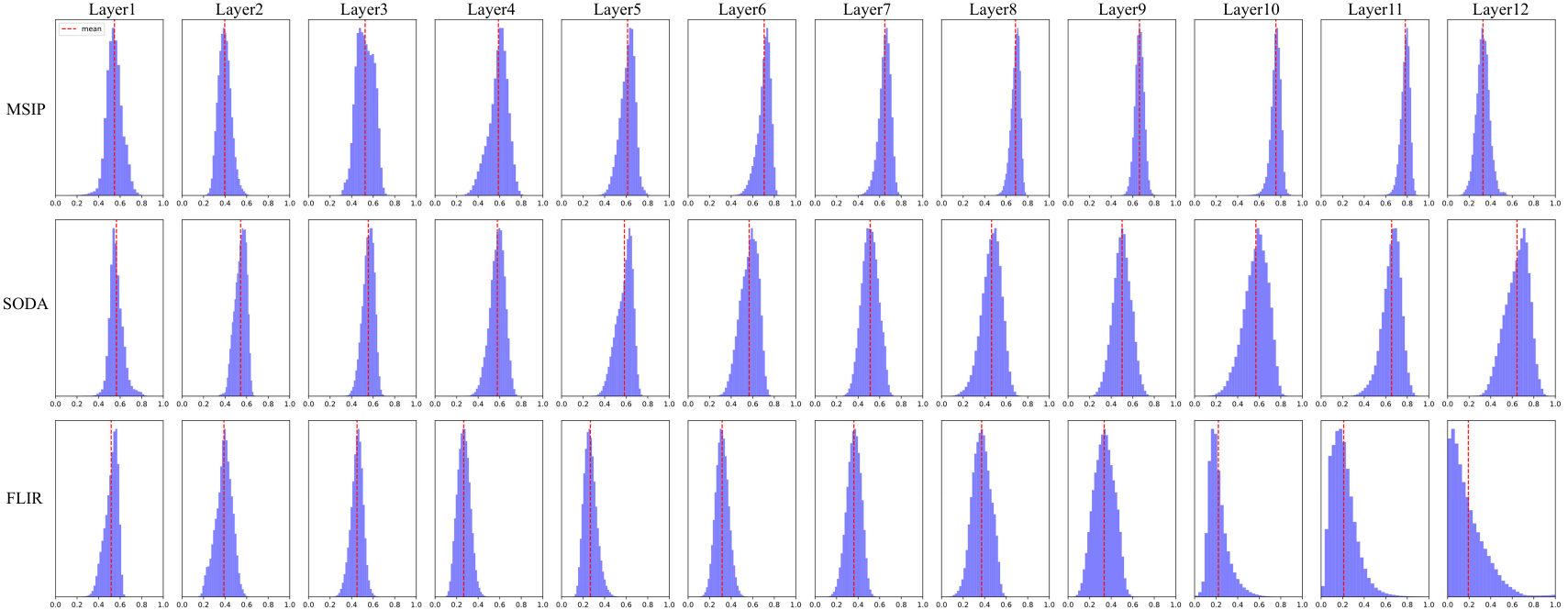}
  \vspace{-6mm}
  \caption{\textbf{Histograms of the scale factors of different patches in each layer for different tasks.} The three rows display the inference results for the pre-trained model on the MSIP dataset, the segmentation model fine-tuned using SODA on the SODA dataset, and the detection model fine-tuned using FLIR on the FLIR dataset, respectively. For each dataset, 100 images are randomly selected.} 
  \label{fig:scale_dis}
  \vspace{-4mm}
\end{figure*}

\subsection{Pre-training}
\label{app:pretrain}
All experiments are conducted using the PyTorch toolkit \cite{paszke2019pytorch} on 4 NVIDIA RTX 3090 GPUs. The weights of all models pre-trained on ImageNet are obtained from MMSelfSup \cite{mmselfsup2021}. The default setting is shown in \cref{tab:setting_pretrain}. We use the linear \textit{learning rate} scaling rule: $lr=base\_lr \times$ batchsize / 256, following MAE \cite{mae}. The decay rate of the learning rate in \textit{full pre-training from IN1K} is 0.7.

\subsection{Fine-tuning}
\label{app:finetune}

\noindent \textbf{Object detection.} Following ViTDet \cite{li2022vitdet}, the Detectron2 \cite{wu2019detectron2} is used to train the detection model for 12 epochs. The default setting is given in \cref{tab:setting_det}. Moreover, the learning rate is reduced to one-tenth of its original value at the 9th and 11th epochs.

\noindent \textbf{Semantic segmentation.} The segmentation model is trained for 100 epochs using MMSegmentation \cite{mmseg2020}, following BeiT \cite{bao2021beit}. The default setting is in listed \cref{tab:setting_seg}. The mIoU includes the background class, following \cite{soda}.

\section{Additional ablation studies}
\label{app:experiments}
\noindent \textbf{Random roi cropping.} \Cref{tab:large_ratio} presents ablation experiment results on the components of random RoI cropping. When random translation and scaling are not applied to the RoIs (\cref{tab:large_ratio} (a)), the performance is slightly lower compared to scenarios where these randomizations are employed (the 1st entry of \cref{tab:large_ratio} (b)). This suggests that random translation and scaling contribute to the diversity of RoIs, thus improving performance. Restricting the size of RoIs to ensure a certain ratio of high-resolution images as input enhances performance (the 2nd and 3rd entries of \cref{tab:large_ratio} (b)). However, as the \textit{large ratio} increases (the last entry of \cref{tab:large_ratio} (b)), the random RoI cropping gradually resembles the random resized cropping (see \cref{alg:rrc}), leading to a decline in performance.

\noindent \textbf{Amount of pre-training data.} The impact of using varying proportions of images for pre-training is explored in \cref{tab:ablation_data}. When using only 1\% of the images from MSIP, the performance of the model is only comparable to not pre-training the patchwise-scale adapter (the 1st entry of \cref{tab:ablation_data}), indicating that pre-training with a very limited amount of data does not have much impact. However, increasing the ratio to 10\% results in a noticeable improvement in the model's performance. As the proportion continues to grow, the model exhibits an overall upward trend in performance on downstream tasks. Therefore, it is believed that using larger datasets containing a wider variety of scenes will further enhance the performance of pre-trained models.

\begin{table}[t]
    \centering
    \setlength{\tabcolsep}{1.5mm}{
    \footnotesize
    \begin{tabular}{c|c|c|c|cc}
        \hline
        &  \multirow{2}{*}{Method} & \multirow{2}{*}{Large ratio} & \multirow{2}{*}{SODA} & \multicolumn{2}{c}{MFNet}\\
        & &  & & val & test \\
         \hline
        (a) & RoI cropping & - & 68.20 &   45.41  & 48.63 \\
         \hline
        \multirow{4}{*}{(b)} & \multirow{4}{*}{Random RoI cropping} & 0\%    & 68.26  & 46.32 &  48.68 \\
        & &  15\% & \graybase{\textbf{68.41}} & \graybase{\textbf{46.89}}  & \graybase{48.82} \\
        & &  30\% & 68.26  & 46.51 & \textbf{49.00} \\
        & &  50\% &  68.06  & 45.87 &  48.47 \\ 
         \hline 
    \end{tabular}}
    \vspace{-2mm}
    \caption{\textbf{Ablations for components of random RoI cropping.} The \textit{large ratio} denotes the probability of $LargeRoI$ being true in random RoI cropping. RoI cropping represents that random translation and scaling are removed. Random RoI cropping with \textit{large ratio} being 15\% performs best.}
    \label{tab:large_ratio}
    \vspace{-2mm}
\end{table}

\begin{table}[t]
    \centering
    \setlength{\tabcolsep}{1.5mm}{
    \footnotesize
    \begin{tabular}{cc|l|ll}
        \hline
        \multirow{2}{*}{Ratio} & \multirow{2}{*}{\#Images} & \multicolumn{1}{c|}{\multirow{2}{*}{SODA}} & \multicolumn{2}{c}{MFNet}\\
         &  &   & \multicolumn{1}{c}{val} & \multicolumn{1}{c}{test} \\
        \hline
        0\% & 0 & 67.90 & 45.30 & 47.28 \\
        1\% & 1,788 & 67.85\textcolor{gray}{(-0.05)} & 45.55\textcolor{blue}{(+0.25)} & 47.10\textcolor{gray}{(-0.18)} \\
        10\% & 17,876 & 68.05\textcolor{blue}{(+0.15)} & 45.87\textcolor{blue}{(+0.57)} & 48.28\textcolor{blue}{(+1.00)} \\
        30\% & 53,627 & 68.39\textcolor{blue}{(+0.49)} & 45.91\textcolor{blue}{(+0.61)} & 48.77\textcolor{blue}{(+1.49)} \\
        50\% & 89,378 & 68.24\textcolor{blue}{(+0.34)} & 46.13\textcolor{blue}{(+0.83)} & 48.68\textcolor{blue}{(+1.40)}  \\
        100\% & 178,756 & \graybase{\textbf{68.41}\textcolor{blue}{(+0.51)}} & \graybase{\textbf{46.89}\textcolor{blue}{(+1.59)}} & \graybase{\textbf{48.82}\textcolor{blue}{(+1.54)}}  \\
        \hline
    \end{tabular}}
    \vspace{-2mm}
    \caption{\textbf{Ablations for the amount of pre-training data.} The images are uniformly sampled from the MSIP dataset. A ratio of 0\% indicates that the patchwise-scale adapter is directly used for downstream tasks without pre-training. Leveraging a larger volume of data can lead to improved performance.}
    \label{tab:ablation_data}
    \vspace{-5mm}
\end{table}

\section{Additional visualizations}
\label{app:visualization}

The histograms of the scale factors for each layer in different tasks are illustrated in \cref{fig:scale_dis}. Firstly, for the same task, the mean values of the distributions across different layers are inconsistent (each row in \cref{fig:scale_dis}), indicating varying degrees of domain-specific feature requirements across different layers. Secondly, for different tasks, the mean values are also not the same for the same layer (each column in \cref{fig:scale_dis}), suggesting that the importance of domain-specific features varies across different tasks. The two observations validate the discussion in \cref{subsec:limitations} that the scale factors among different layers in various tasks should be distinct. Thirdly, in deeper layers, there is a larger distribution difference of scale factors between pre-training and downstream tasks. Moreover, the adjustment direction in SODA and FLIR differs. This suggests that layers more relevant to the task (deeper layers) require more task-specific adjustment, which can be provided by the patchwise-scale adapter. This demonstrates the strong applicability to different tasks of the patchwise-scale adapter, mitigating the limitations of employing adapters in pre-training.

\end{document}